# Multi-Region Ensemble Convolutional Neural Network for Facial Expression Recognition

Yingruo Fan, Jacqueline C.K. Lam and Victor O.K. Li

Department of Electrical and Electronic Engineering,
The University of Hong Kong,
Pokfulam, Hong Kong
`yrfan@hku.hk,{jcklam,vli}@eee.hku.hk`

**Abstract.** Facial expressions play an important role in conveying the emotional states of human beings. Recently, deep learning approaches have been applied to image recognition field due to the discriminative power of Convolutional Neural Network (CNN). In this paper, we first propose a novel Multi-Region Ensemble CNN (MRE-CNN) framework for facial expression recognition, which aims to enhance the learning power of CNN models by capturing both the global and the local features from multiple human face sub-regions. Second, the weighted prediction scores from each sub-network are aggregated to produce the final prediction of high accuracy. Third, we investigate the effects of different sub-regions of the whole face on facial expression recognition. Our proposed method is evaluated based on two well-known publicly available facial expression databases: AFEW 7.0 and RAF-DB, and has been shown to achieve the state-of-the-art recognition accuracy.

## 1 Introduction

Facial expression recognition (FER) has many practical applications such as treatment of depression, customer satisfaction measurement, fatigue surveillance and Human Robot Interaction (HRI) systems. Ekman et al. [2] defined a set of prototypical facial expressions (e.g. anger, disgust, fear, happiness, sadness, and surprise). Since Convolutional Neural Network (CNN) has already proved its excellence in many image recognition tasks, we expect that it can show better results than already existing machine learning methods in facial expression prediction problems. A well-designed CNN trained on millions of images can parameterize a hierarchy of filters, which capture both low-level generic features and high-level semantic features. Moreover, current Graphics Processing Units (GPUs) expedite the training process of deep neural networks to tackle big-data problems. However, unlike large scale visual object recognition databases such as ImageNet [10], existing facial expression recognition databases do not have sufficient training data, resulting in overfitting problems.

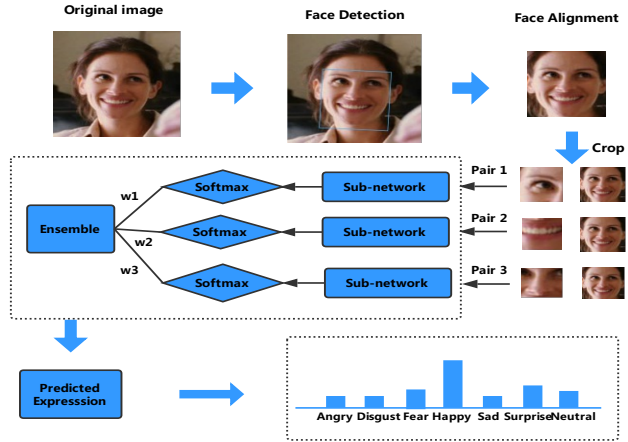

**Fig. 1.** An overview of our approach: Multi-Region Ensemble CNN (MRE-CNN) framework.

CNN approaches topped three slots in the 2014 ImageNet challenge for object recognition task, with the VGGNet [11] architecture achieving a remarkably low error rate. With a review of previous CNNs, AlexNet [5] demonstrated the effectiveness of CNN by introducing convolutional layers followed by Max-pooling layers and Rectified Linear Units (ReLUs). AlexNet significantly outperformed the runner-up with a top-5 error rate of 15.3% in the 2012 ImageNet challenge. In our proposed framework, one of the network structures is based on AlexNet and the other one VGG-16 is a deeper network based on VGGNet [11].

The goal of automatic FER is to classify faces in static images or dynamic image sequences as one of the six basic emotions. However, it is still a challenging problem due to head pose, image resolution, deformations, and illumination variations. This paper is the first attempt to exploit the local characteristics of different parts of the face by constructing different sub-networks. Our main contributions are three-fold and can be summarized as follows:

- A novel Multi-Region Ensemble CNN framework is proposed for facial expression recognition, which takes full advantage of both global information and local characteristics of the whole face.
- Based on the weighted sum operation of the prediction scores from each sub-network, the final recognition rate can be improved compared to the original single network.
- Our MRE-CNN framework achieves a very appealing performance and outperforms some state-of-the-art facial expression methods on AFEW 7.0 Database [1] and RAF-DB [6].

## 2   Related Work

Several studies have proposed different architectures of CNN in terms of FER problems. Hu et al. [3] integrated a new learning block named Supervised Scoring Ensemble (SSE) into their CNN model to improve the prediction accuracy. This has inspired us to incorporate other well-designed learning strategies to existing mainstream networks bring about accuracy gains. [8] followed a transfer learning approach for deep CNNs by utilizing a two-stage supervised fine-tuning on the pre-trained network based on the generic ImageNet [10] datasets. This implies that we can narrow down the overfitting problems due to limited expressions data via transfer learning. In [7], inception layers and the network-in-network theory were applied to solve the FER problem, which focuses on the network architecture. However, most of the previous methods have processed the entire facial region as the input of their CNN models, paying less attention to the sub-regions of human faces. To our knowledge, few works have been done by directly cropping the sub-regions of facial images as the input of CNN in FER. In this paper, each sub-network in our MRE-CNN framework will process a pair of facial regions, including a whole-region image and a sub-region image.

## 3   The Proposed Method

The overview of our proposed MRE-CNN framework is shown in Figure 1. We will start with the data preparation, and then describe the detailed construction for our MRE-CNN framework.

### 3.1   Data Pre-processing

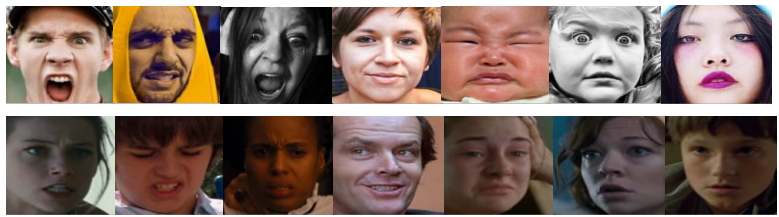

**Fig. 2.** The first row displays cropped faces extracted from images in RAF-DB, and the second row represents faces sampled across video clips in AFEW 7.0.

**Datasets** Recently, Real-world Affective Faces Database (RAF-DB)[1], which contains about 30000 real-world facial images from thousands of individuals, is released to encourage more research on real-world expressions. The images (12271 training samples and 3068 testing samples) in RAF-DB were downloaded from Flickr, after which humans were asked to pick out images related with the six basic emotions, plus the neutral emotion. The other database, Acted Facial Expressions in the Wild (AFEW 7.0) [1], was established for the 2017 Emotion Recognition in the Wild Challenge [2](EmotiW). AFEW 7.0 consists of training (773), validation (383) and test (653) video clips, where samples are labeled with seven expressions: angry, disgust, fear, happy, sad, surprise and neutral.

**Face Detection and Alignment** For each video clip in AFEW 7.0, we sample at 3-10 frames that have clear faces with an adaptive frame interval. To extract and align faces both from original images in RAF-DB and frames of videos in AFEW 7.0, we use a C++ library, Dlib[3] face detector to locate the 68 facial landmarks. As shown in Figure 3, based on the coordinates of localized landmarks, aligned and cropped whole-region and sub-regions of the face image can be generated in a uniform template with a affine transformation. In this stage, we align and crop regions of the left eye, regions of the nose, regions of the mouth, as well as the whole face. Then three pairs of images are all resized into 224×224 pixels.

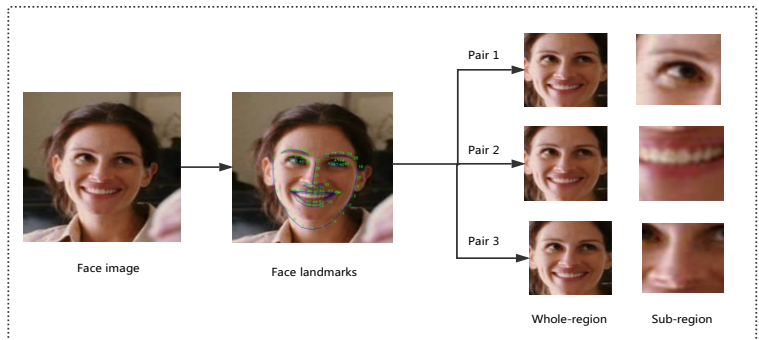

**Fig. 3.** The processing of the cropped whole-region and sub-regions of the facial image.

### 3.2 Multi-Region Ensemble Convolutional Neural Network

Our framework is illustrated in Figure 1. We take three significant sub-regions of the human face into account: the left-eye, the nose and the mouth. Each particular sub-region will be accompanied by its corresponding whole facial image, forming a double input subnetwork in Multi-Region Ensemble CNN (MRE-CNN) framework. Afterwards, based on the weighted sum operation of three prediction scores from each sub-network, we get a final accurate prediction.

---

[1] http://www.whdeng.cn/RAF/model1.html
[2] https://sites.google.com/site/emotiwchallenge/
[3] dlib.net

Particularly, to encourage intra-class compactness and inter-class separability, each subnet adopts the softmax loss function which is given by

$$Loss(\theta) = -\frac{1}{m}\left[\sum_{i=1}^{m}\sum_{j=1}^{k} l\{y^{(i)} = j\}\right] \log \frac{e^{\theta_j^T x^{(i)}}}{\sum_{l=1}^{k} e^{\theta_l^T x^{(i)}}}, \quad (1)$$

where $x^{(i)}$ denotes the features of the i-th sample, taken from the final hidden layer before the softmax layer, $m$ is the number of training data, and $k$ is the number of classes. We define the i-th input feature $x^{(i)} \in \mathcal{R}^d$ with the predicted label $y^{(i)}$. $\theta$ is the parameter matrix of the softmax function $Loss(\theta)$. Here $l\{\cdot\}$ means $l\{a\ true\ statement\} = 1$ or $l\{a\ false\ statement\} = 0$.

**Data Augmentation** Despite the training size of RAF-DB, it is still insufficient for training a designed deep network. Therefore, we utilize both offline data augmentation and on-the-fly data augmentation techniques. The number of training samples increases fifteen-fold after introducing methods including image rotation, image flips and Gaussian distribution random perturbations. Besides, on-the-fly data augmentation is embedded in the deep learning framework, Caffe [4], by randomly cropping the input images and then flipping them horizontally.

### 3.3 The Sub-networks in MRE-CNN Framework

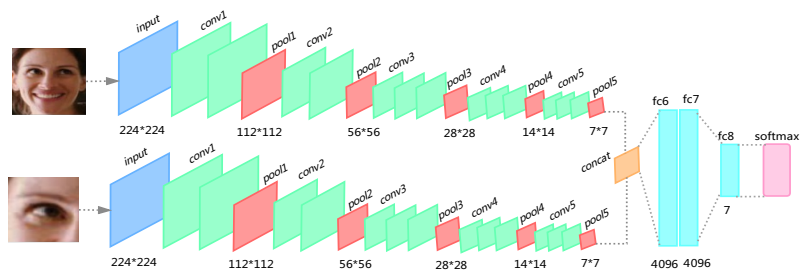

**Fig. 4.** The VGG-16 sub-network architecture in MRE-CNN framework.

As Figure 4 shows, we adopt 13 convolutional layers and 5 max pooling layers and concatenate the outputs from two pool5 layers before going through the first fully connected layer. The final softmax layer gives the prediction scores. When employing VGG-16 [11], we fine-tune the pre-trained model with the training set of AFEW 7.0 and RAF-DB, respectively, in the following experiments.

To validate the proposed MRE-CNN framework, our modified AlexNet architecture do not use any pre-trained models during its training process. For AlexNet sub-network, we use 5 convolutional layers and 3 max pooling layers, the same as in the traditional CNN architecture. Different from the original AlexNet, the last two fully connected layers have 64 outputs and 7 outputs, respectively, making it possible to retrain a deep network with limited data. The following experiment results indicate its effectiveness in the MRE-CNN framework structure, despite its simplified network architecture.

Finally, we combine the three predictions from three sub-networks by conducting the weighted sum operation. The predicted emotion $P_{MRE-CNN}$ is defined as

$$P_{MRE-CNN} = \sum_{n=1}^{z} \alpha_n \sum_{i=1}^{m} \frac{1}{\sum_{l=1}^{k} e^{\theta_l^T x^{(i)}}} \begin{pmatrix} e^{\theta_1^T x^{(i)}} \\ ... \\ e^{\theta_k^T x^{(i)}} \end{pmatrix}, \qquad (2)$$

where $\alpha$ denotes the weight for a single sub-network and $z$ is equal to 3 as we utilize three sub-networks. Other parameters are the same as those in Equation 1.

## 4 Experiments

### 4.1 Experimental Setup

All training and testing processes were performed on NVIDIA GeForce GTX 1080Ti 11G GPUs. We developed our models in the deep learning framework Caffe [4]. On the Ubuntu linux system equipped with NVIDIA GPUs, training a single model in MRE-CNN took 4-6 hours depending on the architecture of the sub-network.

### 4.2 Implementation Details

In data augmentation stage, we augment the set of training images in RAF-DB and frames in AFEW 7.0 by flipping, rotating each with $\pm 4°$ and $\pm 6°$, and adding Gaussian white noises with variances of 0.001, 0.01 and 0.015. We then train our VGG-16 sub-networks for 20k iterations with the following parameters: learning rate 0.0001-0.0005, weight decay 0.0001, momentum 0.9, batch size 16 and linear learning rate decay in stochastic gradient descent (SGD) optimizer. For AlexNet sub-networks, we train them for 30k iterations with the batch size of 64 and the learning rate begins from 0.001. In the ensemble prediction stage, the specific weights of MRE-CNN (VGG-16 Sub-network) are 4/7 (left-eye weight), 2/7 (mouth weight) and 1/7 (nose weight) and those of MRE-CNN (AlexNet Sub-network) are 2/5 (left-eye weight), 2/5 (mouth weight) and 1/5 (nose weight), respectively.

### 4.3 Results on RAF-DB

RAF-DB is split into a training set and a test set with the idea of five-fold cross-validation and we performed the 7-class basic expression classification benchmark experiment. In the RAF-DB test protocol, the ultimate metric is the mean diagonal value of the confusion matrix rather than the accuracy due to imbalanced distribution in expressions. In this experiment, we directly train our deep learning models with our processed training samples from RAF-DB, without using other databases. In details, after filtering the non-detected face images and applying data augmentation techniques, 95465 cropped face images are generated, accompanied by left-eye images, mouth images and nose images.

**Table 1.** Confusion matrix for RAF-DB based on MRE-CNN (VGG-16 Sub-network). The term Real represents the true labels (0=Angry, 1=Disgust, 2=Fear, 3=Happy, 4=Sad, 5=Surprise, 6=Neutral) and Pred represents the predicted value.

| Real \ Pred | 0 | 1 | 2 | 3 | 4 | 5 | 6 |
|---|---|---|---|---|---|---|---|
| 0 | 0.0088 | 0.0632 | 0.0000 | 0.0221 | 0.0706 | 0.0338 | **0.8015** |
| 1 | 0.0213 | 0.0182 | 0.0334 | 0.0030 | 0.0122 | **0.8602** | 0.0517 |
| 2 | 0.0209 | 0.0565 | 0.0084 | 0.0167 | **0.7992** | 0.0105 | 0.0879 |
| 3 | 0.0110 | 0.0211 | 0.0051 | **0.8878** | 0.0127 | 0.0110 | 0.0515 |
| 4 | 0.0811 | 0.0000 | **0.6081** | 0.0270 | 0.0676 | 0.1757 | 0.0405 |
| 5 | 0.1125 | **0.5750** | 0.0063 | 0.0083 | 0.0750 | 0.0187 | 0.1313 |
| 6 | **0.8395** | 0.0802 | 0.0185 | 0.0185 | 0.0123 | 0.0062 | 0.0247 |

Analyzing the confusion matrix based on MRE-CNN (VGG-16 Sub-network) in Table 1, our proposed model performs well when classifying happy, surprise and angry emotions, with accuracy of 88.78%, 86.02%, 83.95%, respectively. For comparison, in Table 2 we show the results of the trained DCNN models followed by different classifiers which are proposed in [6]. We find that our proposed MRE-CNN (VGG-16) framework outperforms all of the existing state-of-the-art methods evaluated on RAF-DB. In addition, the MRE-CNN (AlexNet) framework also achieves a very appealing performance although we retrain the AlexNet sub-networks with limited data.

**Table 2.** Performance of different methods on RAF-DB (The metric is the mean diagonal value of the confusion matrix).

|  | Angry | Disgust | Fear | Happy | Sad | Surprise | Neutral | Ave. |
|---|---|---|---|---|---|---|---|---|
| DLP+SVM[6] | 71.60 | 52.15 | 62.16 | 92.83 | 80.13 | 81.16 | 80.29 | 74.20 |
| DLP+LDA[6] | 77.51 | 55.41 | 52.50 | 90.21 | 73.64 | 74.07 | 73.53 | 70.98 |
| Alex+SVM[6] | 58.64 | 21.87 | 39.19 | 86.16 | 60.88 | 62.31 | 60.15 | 55.60 |
| Alex+LDA[6] | 43.83 | 27.50 | 37.84 | 75.78 | 39.33 | 61.70 | 48.53 | 47.79 |
| VGG+SVM[6] | 68.52 | 27.50 | 35.13 | 85.32 | 64.85 | 66.32 | 59.88 | 58.22 |
| VGG+LDA[6] | 66.05 | 25.00 | 37.84 | 73.08 | 51.46 | 53.49 | 47.21 | 50.59 |
| VGG-FACE | 82.19 | 56.62 | 55.41 | 86.38 | 79.52 | 83.93 | 71.18 | 73.60 |
| Ours(AlexNet) | 77.78 | 65.62 | 58.11 | 87.75 | 75.73 | 81.16 | 77.21 | 74.78 |
| Ours(VGG-16) | 83.95 | 57.50 | 60.81 | 88.78 | 79.92 | 86.02 | 80.15 | **76.73** |

Furthermore, we separated the sub-network modules from MRE-CNN framework and demonstrated their individual results on the test set of RAF-DB. Results can be viewed in Table 3. The result of the first row shows the average accuracy of Face+Left-Eye while applying VGG-16 sub-network in MRE-CNN framework, and they are higher than that of Face+Mouth. Thus we assign higher weights to Face+LeftEye subnet when combining the three predictions with an appropriate ensemble method. Face+Nose subnet is slightly less effective, probably due to less information related to emotions; Nevertheless, it is still superior to the VGG-FACE model given in Table 2 with only the whole face region as input.

**Table 3.** Sub-region Comparison(The metric is the mean diagonal value of the confusion matrix).

| Architecture | Average |
| --- | --- |
| Face+LeftEye (Single VGG-16 sub-network) | 76.52 |
| Face+Nose (Single VGG-16 sub-network) | 75.64 |
| Face+Mouth (Single VGG-16 sub-network) | 76.13 |
| **Our MRE-CNN(VGG-16)** | **76.73** |

**Table 4.** Comparisons with the state-of-the-art methods on AFEW 7.0(The metric is the average accuracy of all validation videos).

| Architecture | Training data | Average |
| --- | --- | --- |
| C3D [9] | 16 frames for each video | 35.20 |
| ResNet-LSTM[9] | 16 frames for each video | 46.70 |
| VGG-LSTM[9] | 16 frames for each video | 47.40 |
| Trajectory+SVM[13] | 30 frames for each video | 37.37 |
| VGG-BRNN[13] | 40 frames for each video | 44.46 |
| C3D-LSTM[12] | Detected face frames | 43.20 |
| **Our MRE-CNN(AlexNet)** | Detected face frames | 40.11 |
| **Our MRE-CNN(VGG-16)** | Detected face frames | **47.43** |

### 4.4 Results on AFEW 7.0

To validate the performance of our models, we also conduct experiments on the validation set of AFEW 7.0. The task is to assign a single expression label from seven candidate categories to each video clip from the validation set (383 video clips). Note that all our CNN models in MRE-CNN framework are trained on the given training data (773 video clips) only without applying any outside data. Considering the temporally disappearance or occlusion in some videos, we only use detected face frames for training and prediction. In our experiments, the predicted emotion scores of each video are calculated by averaging the scores of all its detected face frames. We can see from

Table 4, for the validation set of AFEW 7.0, our MRE-CNN (VGG-16) framework gets great results which are superior to some state-of-the-art methods.

### 4.5 Discussions

A series of feature maps are shown in Figure. 5 for VGG-16 sub-network in our MRE-CNN framework, which can reflect the differences in the filters of the first three convolutional layers. It can be observed that shallower layer outputs capture more profile information while deeper layer outputs encode the semantic information. Shallower layers can learn rich low-level features that can help refine the irregular features from deeper layers. Furthermore, by combining features from the whole region and sub-regions of the human face, the resulting architecture provides more rich feature maps, which raises the recognition rate for FER problems.

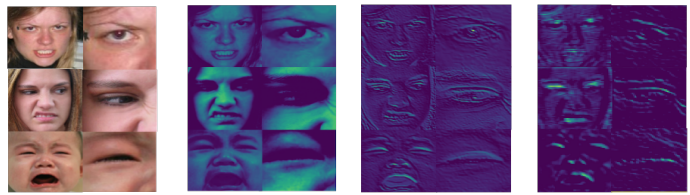

**Fig. 5.** Visualization of the feature maps of the first three convolutional layers for the input image on the left of each row.

Generally, our method explicitly inherits the advantage of information gathered from multiple local regions from face images, acting as a deep feature ensemble with two single CNN architectures, and hence it naturally improves the final predication accuracy. The disadvantage of our approach is that we use grid searching to determine the contribution portions of individual sub-networks, which is relatively computationally expensive. Although facial expression recognition based on face images can achieve promising results, facial expression is only one modality in realistic human behaviors. Combining facial expressions with other modalities, such as audio information, physiological data and thermal infrared images can provide complementary information, further enhancing the robustness of our models. Therefore, it is a promising research direction to incorporate facial expression models with other dimension models into a high-level framework.

## 5 Conclusion

We proposed a novel Multi-Region Ensemble CNN framework in this study, which takes full advantage of the different regions of the whole human face. By assigning different weights to three sub-networks in MRE-CNN, we have combined the predictions of three separate networks. Besides, we have investigated the effects of three different facial regions, each providing different local information. As a result, our MRE-CNN framework has achieved a very appealing performance on RAF-DB and AFEW 7.0, as compared to other state-of-the-art methods.

**Acknowledgements.** This research is supported in part by the Theme-based Research Scheme of the Research Grants Council of Hong Kong, under Grant No. T41-709/17-N.